\documentclass[10pt,a4paper]{article}
\usepackage{graphicx}
\usepackage{subfigure}
\usepackage{amsmath,amsthm}
\usepackage{amssymb,amsfonts}
\usepackage{threeparttable}
\usepackage{multirow}
\usepackage{booktabs}
\usepackage{authblk}
\usepackage{url}
\usepackage{fdsymbol}

\newtheorem{assumption}{\textbf{Assumption}}
\newtheorem{definition}{\textbf{Definition}}

\newtheorem{proposition}{\textbf{Proposition}}

\title{Causal Domain Adaptation with Copula Entropy based Conditional Independence Test}
\author{Jian MA\thanks{Email: majian@hitachi.cn}}
\affil{Hitachi China Research Laboratory}

\begin{document}

\maketitle

\begin{abstract}
\noindent
Domain Adaptation (DA) is a typical problem in machine learning that aims to transfer the model trained on source domain to target domain with different distribution. Causal DA is a special case of DA that solves the problem from the view of causality. It embeds the probabilistic relationships in multiple domains in a larger causal structure network of a system and tries to find the causal source (or intervention) on the system as the reason of distribution drifts of the system states across domains. In this sense, causal DA is transformed as a causal discovery problem that finds invariant representation across domains through the conditional independence between the state variables and observable state of the system given interventions. Testing conditional independence is the corner stone of causal discovery. Recently, a copula entropy based conditional independence test was proposed with a rigorous theory and a non-parametric estimation method. In this paper, we first present a mathemetical model for causal DA problem and then propose a method for causal DA that finds the invariant representation across domains with the copula entropy based conditional independence test. The effectiveness of the method is verified on two simulated data. The power of the proposed method is then demonstrated on two real-world data: adult census income data and gait characteristics data. 
\end{abstract}

\section{Introduction}
Domain Adaptation (DA) is a typical problem of machine learning that gains much attentions recently \cite{Wilson2020,Zhuang2021,Kouw2021}. The problem arises when there are data with different distributions from multiple domains and one want to use the model trained on the data from one domain to another different domain. Since the distributions differs across domains, such direct model transfer may fail with degraded performance in most of time. The aim of the problem is to learn a transferable common model which can be deployed across domains. 

The problem of DA can be widely applied to many different fields, such as medicine \cite{Sun2020}, neuroscience \cite{Prahm2019,Chen2020e}, or social sciences \cite{Budhathoki2021a}. For example, in medicine, different population of the patients with same disease may receive different treatments, and hence each sub-population of a treatment may produce clinical data that differs from that of other sub-populations since body states may be different due to treatments. Researchers may be interested in learning a common knowledge of the disease from the data of different treatments. In social science, when a policy is applied, target population may be divided into many groups in terms of social factors, such as gender, education, culture, incomes, etc. The effect of the policy on those groups may differ due to these factors. Researcher usually study only a particular group and therefore may draw a conclusion that cannot be transferred to other groups directly.

In this work, our motivation is to build a model for fall risk assessment from gait characteristics. Previously, we collected the data of gait characteristics in Timed Up and Go (TUG) test scenarios and build a model that can predict TUG score from gait characteristics for clinical use \cite{ma2020predicting}. However, the model derived from the data of TUG test scenarios cannot be used in daily life scenarios because the gait characteristics of these two scenarios are different as shown in the previous chapter. This problem can be considered as a DA problem with TUG test and daily life scenarios as two domains.

There are a lot of research contributing to DA problem from different aspects. Please refer to \cite{Wilson2020,Zhuang2021,Kouw2021,farahani2020brief} for the reviews of research on DA. 

Causal DA is a special case of DA that tackle the problem from the view of causality \cite{Mooij2020}. It sees multiple domains as parts of a system and distribution drifts as effect of a causal source outside of these domains. In this sense, one embeds the probabilistic relationships in multiple domains into a larger causal network of the system and study the causal source of the change of the underlying distribution across domains. The problem is then transformed as a causal discovery problem that can be solved with many existing methods.

Conditional Independence (CI) test is the basic building block of causal discovery methods. There are many non-parametric methods for such testing, such as conditional distance correlation \cite{wang2015conditional}, kernel-based conditional independence tests \cite{zhang2011kernel}, conditional dependence coefficient \cite{Azadkia2021}, generalised covariance measure \cite{shah2020the}, and the basic and kernel partial correlation \cite{huang2020kernel}, etc. 

Copula Entropy (CE) is a recently introduced concept for statistical independence measurement \cite{ma2008}. It is an ideal tool for statistical independence testing with several axiomatic properties, such as multivariate, symmetric, non-positive (0 iff independent), invariant to monotonic transformation, and equivalent to correlation coefficient in Gaussian cases. It is proved to be related to Transfer Entropy (TE), which is essentially a measure for conditional independence and can be represented with only CE \cite{ma2021estimating}. This CE-based CI measure is advantageous over the other measures in theory since it is based on the rigorously defined CE theory and therefore mathematically sound. Like several other measures, it is distribution-free.

The non-parametric method for estimating CE was also proposed in \cite{ma2008}, which is rank-based and essentially to estimate the entropy of normalized ranks. According to the CE-based representation of TE/CI, a non-parametric method for estimating TE or testing CI was proposed in \cite{ma2021estimating}. It is based on the CE estimation method and has two simple and elegant steps that can be easily implemented in practice.

In this work, we propose a method for solving causal DA as a causal discovery problem with the CE-based CI test. We will first present the basic theory of causal DA. Then we will propose a method that transforms the causal DA problem into a CI test problem and solve it with the CE-based CI estimation method. The proposed method will be evaluated on both simulated and real-world data.

\section{Related work}
Finding a invariant/transferable representation between source and target domains is an intuitive idea for DA. Some of the work try to use statistical dependence measures to derive such invariant dependence structure across domains. Sun et al. \cite{Sun2016} proposed to use the simple second-order statistics for unsupervised DA. Chen et al. \cite{Chen2020} proposed to minimizing the discrepancy of feature distribution between domains with high-order statistics (mainly third and fourth-order). Mutual Information (MI) is a measure of statistical independence in information theory. There are several works on applying information-theoretical measures to domain adaptations. Zhao et al. \cite{zhao2021domain} proposed a framework for DA which learns the features that can transfer between domains by maximizing the MI between features of the same class in both source and target domains. Chen et al. \cite{chen2022preserving} also used MI to learn domain-invariant representations. At the same time, they also tried to mitigate domain divergence by maximizing the MI between the target domain and its private characteristics. Kernel-based dependence measure is an important dependence measure in machine learning, and is also considered in unsupervised DA. Long et al. \cite{Long2015} used maximum mean discrepancy to improve the generalization performance of deep neural network on novel task by embedding task-specific network structure into kernel space and then minimizing the discrepancy between domains.

Another line of the related work is to consider DA as a causal learning problem. Zhang et al. \cite{zhang2015multi} assumed the class is the cause of the features in multiple domains and reconstruct this causal relationship in the target domain based on those in source domain. Magliacane et al. \cite{magliacane2018domain} proposed to introduce contextual intervention on systems to explain the distribution drifts between source and target domains and then presented an approach that exploits causal inference to solve the problem. Mooij et al. \cite{Mooij2020} proposed a framework modelling the DA problem as a structural causal model and suggested using causal discovery algorithm to implement it. Zhang et al. \cite{zhang2020domain} Also proposed a similar idea that tackle DA problem with graphical models. Oberst et al. \cite{Oberst2021} assumed linear causal structure model and proposed a regularization learning algorithm that can balance in-distribution performance and invariance to intervention.

\section{CE based CI test}
Copula theory is a probabilistic theory on representation of multivariate dependence \cite{nelsen2007,joe2014}. According to Sklar's theorem \cite{sklar1959}, any multivariate density function can be represented as a product of its marginals and copula density function (cdf) which represents dependence structure among random variables. 

With copula theory, Ma and Sun \cite{ma2008} defined a new mathematical concept, named Copula Entropy, as follows:
\begin{definition}[Copula Entropy]
Let $\mathbf{X}$ be random variables with marginals $\mathbf{u}$ and copula density function $c$. The CE of $\mathbf{X}$ is defined as
\begin{equation}
	H_c(\mathbf{x})=-\int_{\mathbf{u}}{c(\mathbf{u})\log c(\mathbf{u})d\mathbf{u}}.
\label{eq:ce}
\end{equation}	
\end{definition}
They also proved that CE is equivalent to MI in information theory \cite{infobook}. CE has several ideal properties, such as multivariate, symmetric, invariant to monotonic transformation, non-positive (0 iff independent), and equivalent to correlation coefficient in Gaussian cases. It is a perfect measure for statistical independence. 

CE has also theoretical relationship with CI. Ma \cite{ma2021estimating} proved that Transfer Entropy (TE) can be represented with only CE. Since TE is essentially conditional MI, an information-theoretical measure of CI, we can also measure CI with only CE as the proposition below.
\begin{proposition}
Given random variables $x,y,z$, the measure $H_{ci}$ of conditional independence between $(x,y)$ given $z$ can be represented as follows:
\begin{equation}
	H_{ci}(x,y,z)=H_c(x,z)+H_c(y,z)-H_c(x,y,z).
\label{eq:ci}
\end{equation}
\label{prop1}
\end{proposition}
Please refer to \cite{ma2021estimating} for the proof of this proposition.

Ma and Sun \cite{ma2008} also proposed a non-parametric method for estimating CE, which composes of two simple steps: 1) estimating empirical cdf; and 2) estimation CE from the estimated empirical cdf. In the first step, the rank statistic is used to derive empirical cdf; in the second step, the famous KSG method \cite{kraskov2004} for estimating entropy is suggested. The proposed estimation method is rank-based and to estimate the entropy of rank statistic essentially. With this estimation method, Ma also proposed a non-parametric method for estimating TE or testing CI by estimating the 3 CE terms according to \eqref{eq:ci} \cite{ma2021estimating}.

In a word, CE provides a unified theoretical framework for testing unconditional and conditional independence with the non-parametric methods for estimating CE/TE/CI.

\section{Motivation Problem}
This work is motivated by the issue we face in developing fall risk assessment method for elderly. Previously, we have developed a method for fall risk assessment which predicts TUG scores from gait characteristics with video analysis and machine learning \cite{ma2020predicting}. The original data used in this research was collected from the TUG test on elderly and the data of gait characteristics extracted from the original data were used to build a predictive model. Now we want to develop a model for automated fall risk assessment in daily life scenarios. However, the previously built model cannot be naively applied to daily life scenarios because gait characteristics are different in these two scenarios. Due to intervention of TUG test, people in daily life tend to have slower gait speed, smaller pace and speed variability, more frequent stride, and smaller acceleration range than in TUG test \cite{ma2022comparison}. We should tackle the distribution drift to develop a model for daily life scenarios. This is a typical DA problem.

\section{Theory}
We generally formalize our causal DA problem in this paper. Let $x_i,i=1,\ldots,n$ be the state variables of a system $S$ of interest. The state variables $x_i$ may be interrelated with each other under the governing underlying mechanism of the system $S$. A subset of state variables $x_i,i=1,\ldots,m,m<n$ of the system $S$ can be intervened by a context variable $I$ which can be continuous or discrete as in our example. By taking different value of $I$, one gets multiple sample data of $x_i$ as multiple domains. The state of the system $S$ represented by this subset can be measured as a observable variable $y$ which is irrelevant to other state variables. We assume that the functional relationship between the subset and the observable variable is invariant to intervention, i.e., the context variable $I$. The problem is that given multi-domain data derived by intervention $I$, to find this invariant function that relates the subset of state variables to the observable variable $y$.

In our problem, the system $S$ is human body and the state variables $x_i$ is gait characteristics that reflect the functional ability of human. Here, the underlying mechanism of body movement is unknown to us, but the gait characteristics are known to be interrelated by definitions. Then the context variable $I$ is the contextual scenarios of movement, such as TUG test or daily life, that will change the distributions of the gait characteristics $x_i$. The TUG score that measures functional ability is the observable variable $y$ that is assumed to relate to a subset of gait characteristics. The aim is to find a subset of $x_i$ to build a model for predicting $y$.

The above problem can be represented as the following mathematical model:
\begin{equation}
	\begin{cases}
	x_i\sim P(\mathbf{x};I)\\
	y=f(x_1,\ldots,m;\theta)
	\end{cases},
\label{eq:model}
\end{equation}
where $i=1,\ldots,n; n>m$ and $I,\theta$ are parameters.

In this model, we introduce a probabilistic function $P$ of $x_i$ with $I$ as parameter. In this way, we assume a common underlying mechanism of state variables across multiple domains. This assumption is reasonable if the data of multiple domains are generated from a same system under different interventions. We also introduce a function $f$ with parameter $\theta$ from $x_i$ to $y$. By this, we assume the function $f$ is invariant across domains (interventions). 

In this model, the following three assumptions for causal discovery in \cite{Mooij2020} should also be held:
\begin{assumption}[exogeneity]
	No state variables $x_i$ causes context variable $I$.
\label{assumption1}
\end{assumption}
\begin{assumption}[randomization]
	No context variable $I$ is confounded with a state variable $c_i$.
\label{assumption2}
\end{assumption}
\begin{assumption}[genericity]
	Context variables $I$ are confounded with each other.
\label{assumption3}
\end{assumption}

The assumption \ref{assumption1} means intervention is from outside of system and not related to state variables. The assumption \ref{assumption2} means the interventions are assigned completely randomly with respect to the system. This assumption depends on experiment design. The assumption \ref{assumption3} is about the relationship between interventions, which is important when interventions are from a complicated context. In many simple cases like ours, the assumption can be easily verified as true.

\section{Method}
With the above model, the problem to find the invariant function across domains is a typical causal discovery problem that can be solved by testing the conditional independence between state variables $x_i$ and observable variable $y$ given context variable $I$. The target subset composed of those state variables that satisfy $x_i \nVbar y \mid I$.

We propose to apply the CE-based CI test to test such relationships directly with the following method. In the method, one only first prepare the data by augmenting state variable and observable variable with context variable and then estimate the CI strength on the augmented data with the CE-based CI estimation method according to \eqref{eq:ci}.

The method has several merits. It is theoretically sound since CE is a rigorously defined mathematical concept and has a proved theoretical relationship with CI. It is model-free since CE-based CI test is model-free. Since CE is a information-theoretical concept and has clear physical meaning, CE-based CI test inherits this merit as well. Therefore, the causal relationships discovered with the method are interpretable to users. It is also easy to implement due to the simplicity of the CE/CI estimation methods.

\section{Simulation Experiments}
We did two simulation experiments to verify the effectiveness of our method. In both simulations, two types of bi-variate distributions with a controlling parameter will be used to generate sample data of two variables $(x_1,x_2)$ and then the generated data will be fed into a probability density function $f$ to derive the outcomes $y$. With the controlling parameter, we can simulate multi-domain data by tuning its value to change the underlying dependence structure of the two variables $(x_1,x_2)$. A third random variable $x_3$ independent of $(x_1,x_2)$ is generated as contrast to the two variables. By a probability density function $f$, we assume there is a common function mapping the controlled variable $(x_1,x_2)$ of multi-domains to the outcomes $y$ of the simulation. We will test the conditional independence between the three variables $x_1,x_2,x_3$ and the outcomes $y$ given the controlling parameter.

\begin{figure}
	\centering
	\subfigure[Experiment 1]{\includegraphics[width=0.48\textwidth]{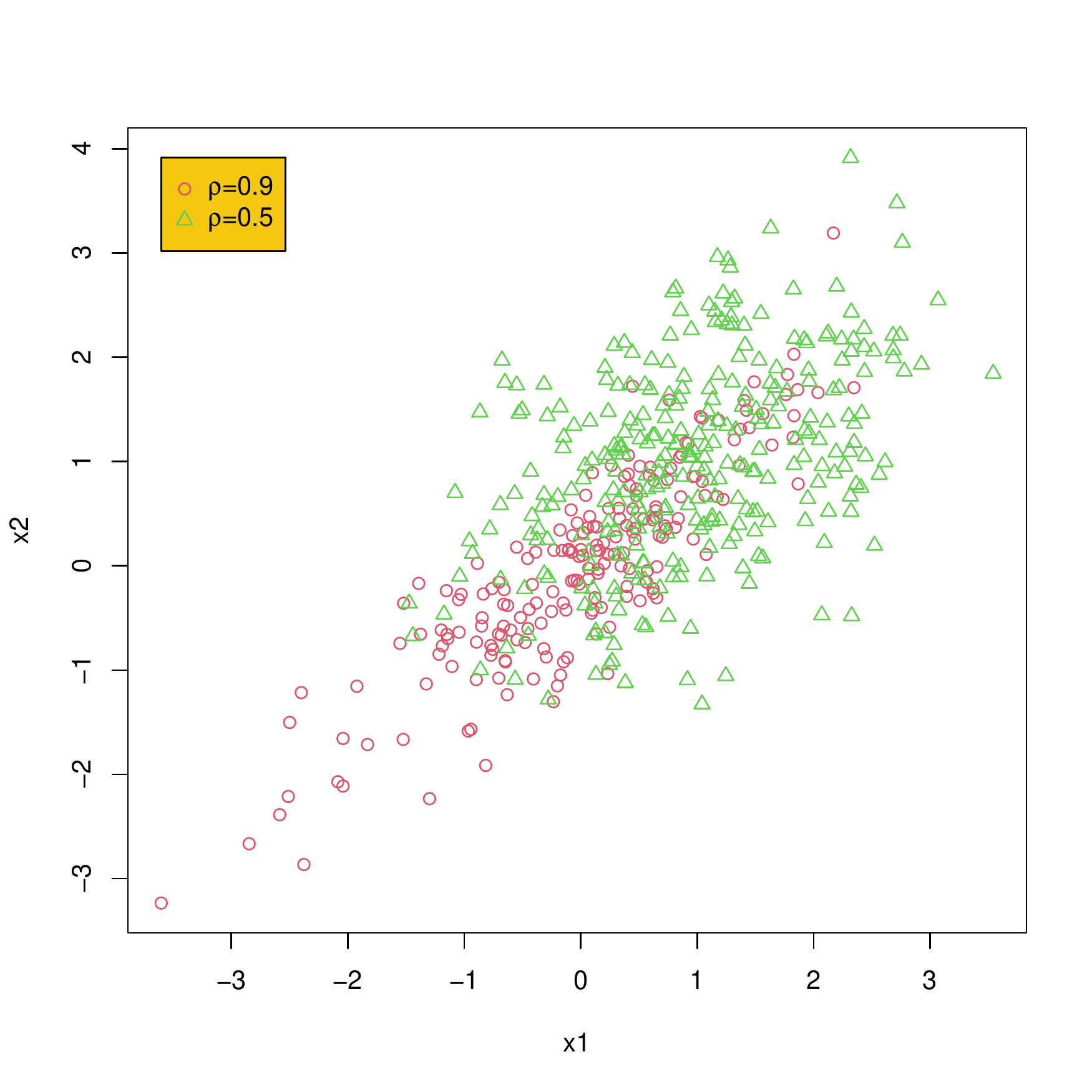}
	\label{fig:sima}}
	\subfigure[Experiment 2]{\includegraphics[width=0.48\textwidth]{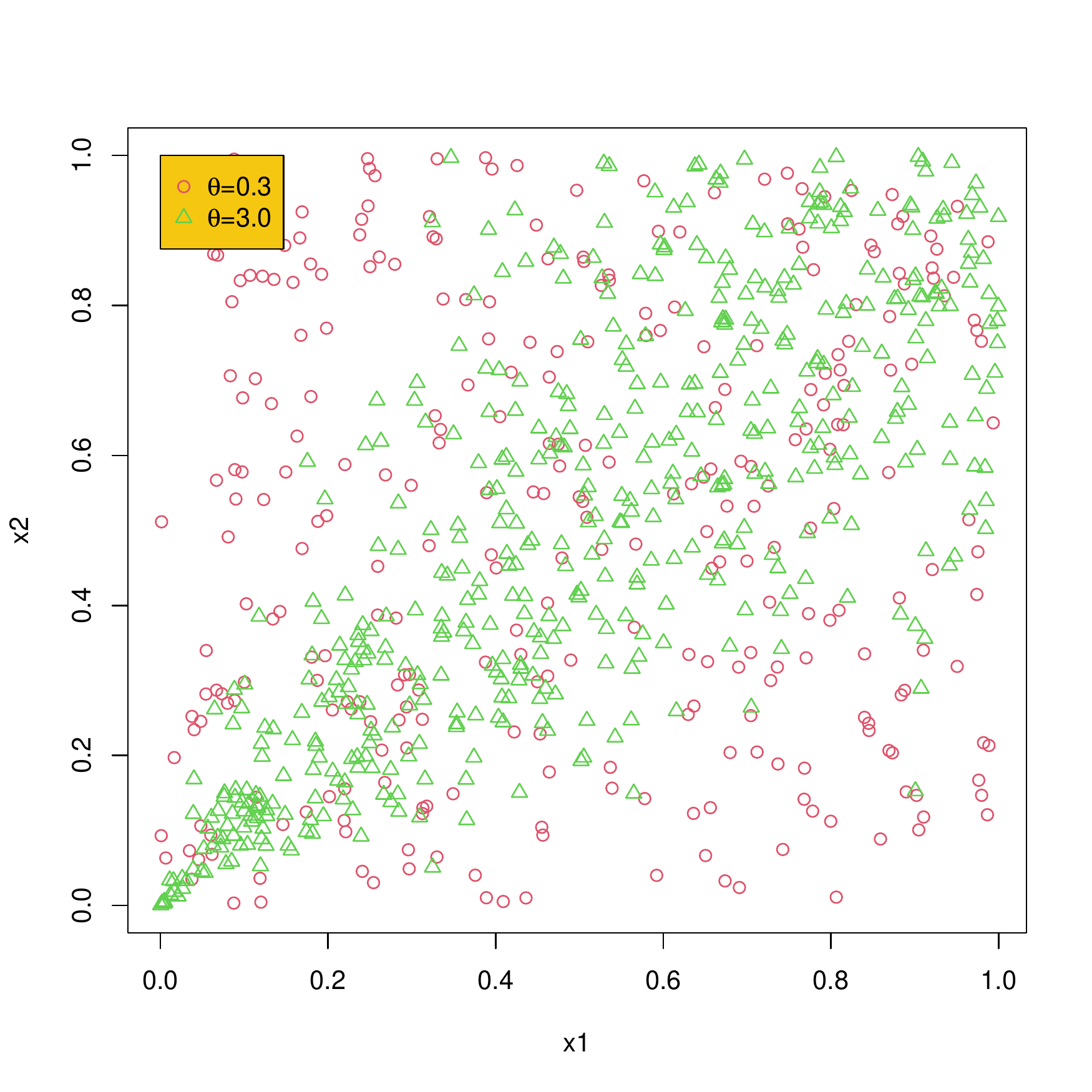}
	\label{fig:simb}}
\caption{Simulated data.}
	\label{fig:simdata}
\end{figure}

In the first experiment, a bi-variate Gaussian distribution $(x_1,x_2)\sim N(\mu,\rho)$ with a zero mean $\mu$ and a covariance $\rho$ was used to generate two samples. In the simulation, two samples of $(x_1,x_2)$ were simulated with the covariance $\rho = 0.5,0.9$ and sample size $N=200,300$ respectively. The mean of the second sample was shifted by $\mu=[1,1]^T$. And therefore, two samples with different underlying dependence structure were derived as shown in Figure \ref{fig:sima}. The two samples were then used as input of a bi-variate normal distribution $y=N(x_1,x_2;0,\rho)$ with a covariance parameter $\rho = 0.8$. A sample ($N=500$) of the third variable $x_3$ was simulated with a normal distribution with mean $\mu = 0$ and variance $\rho = 1$.

In the second experiment, we generated the sample data with a bi-variate non-Gaussian distribution. For non-Gaussianity, a bi-variate clayton Copula function $(x_1,x_2)\sim C_{\theta}^{clayton}$ was used in the experiment:
\begin{equation}
	C_{\theta}^{clayton}(u_1,u_2)=\left(u_1^{-\theta}+u_2^{-\theta}-1\right)^{-\theta^{-1}},
\label{eq:clayton}
\end{equation}
where $\theta \geq 0$ is the parameter that controls the dependence structure between variables. In this simulation, we generated two samples of $(x_1,x_2)$ with $\theta = 0.3,3.0$ and sample size $N=300,500$ respectively. Two samples with different dependence structures were derived as shown in Figure \ref{fig:simb}. Then the generated two samples were used as input of a bi-variate frank copula function to derive the outcome $y\sim C_{\theta}^{frank}$ of the simulation:
\begin{equation}
	C_{\theta}^{frank}(u_1,u_2)=-\frac{1}{\theta}\log\left(1+\frac{(e^{-\theta u_1}-1)(e^{-\theta u_2}-1)}{e^{-\theta}-1}\right),
\end{equation}
where the controlling parameter $\theta \in [-\infty,+\infty]$. In the simulation, $\theta = 0.5$. The sample ($N=800$) of the third variable $x_3$ was simulated with uniform distribution on $[0,1]$. The \textsf{R} package \texttt{copula} \cite{JunYan2007,Hofert2020} was used for the above two copula functions in the simulations.

\begin{figure}
	\centering
	\subfigure[Experiment 1]{\includegraphics[width=0.48\textwidth]{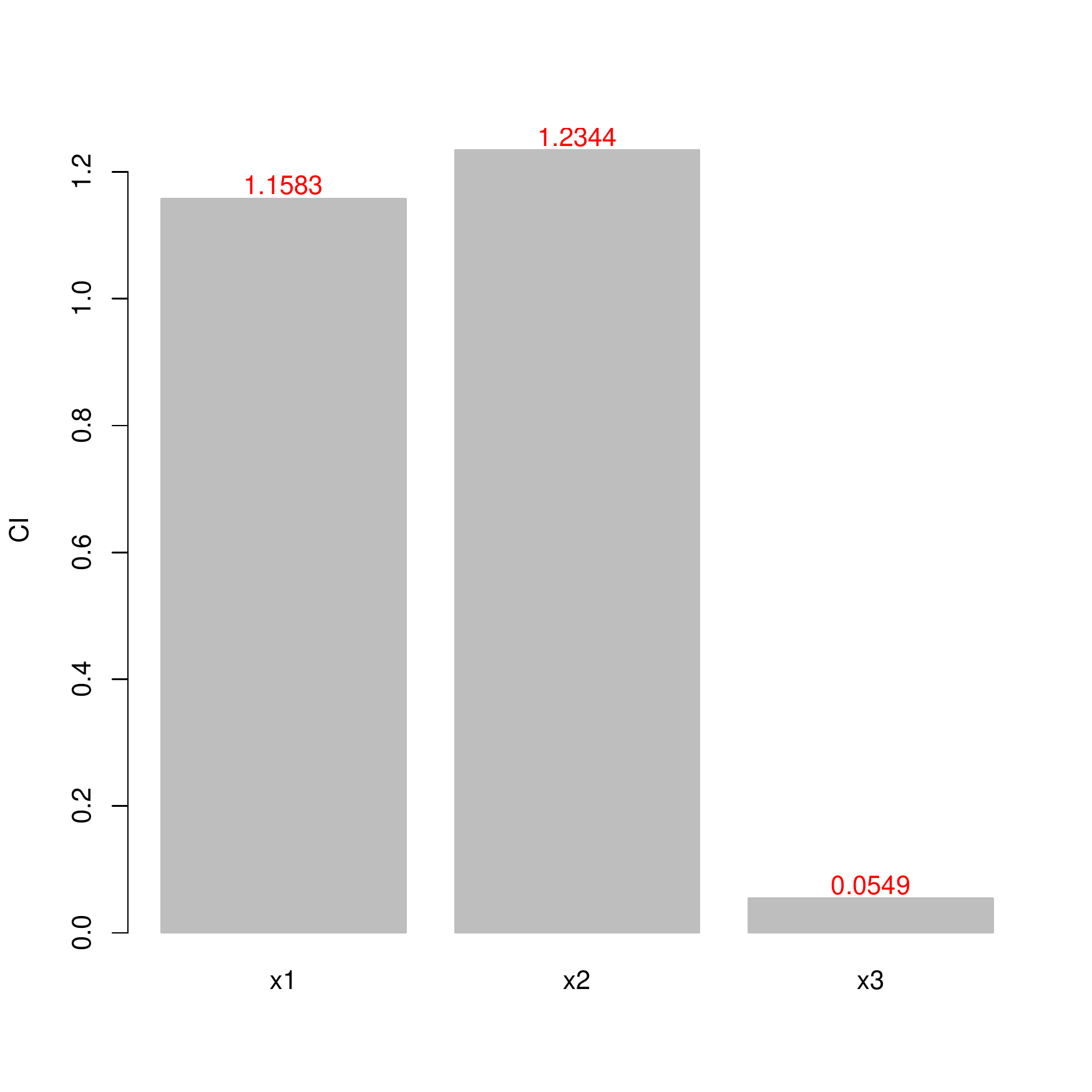}}
	\subfigure[Experiment 2]{\includegraphics[width=0.48\textwidth]{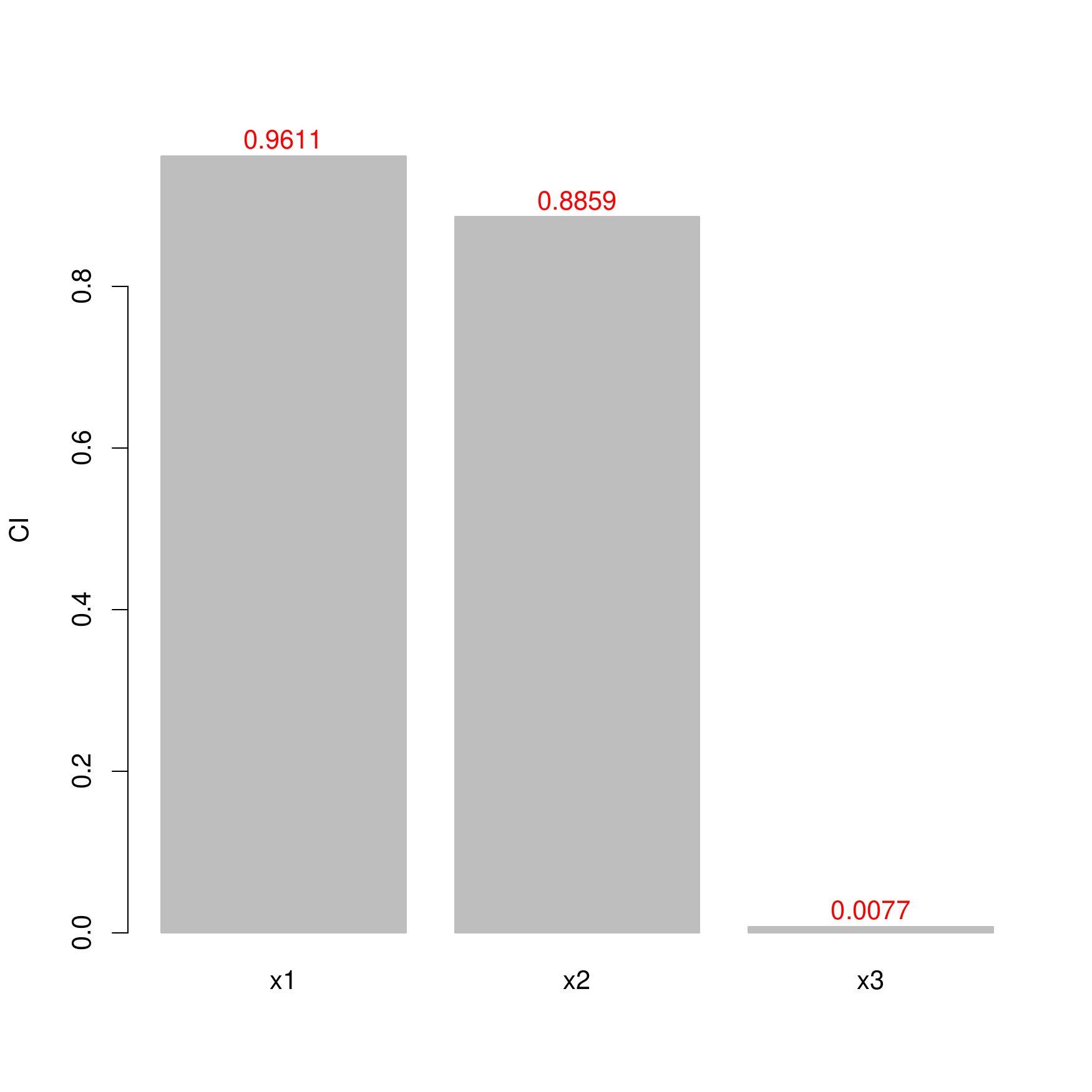}}
	\caption{Results of conditional independence tests of the simulation experiments.}
	\label{fig:simresults}
\end{figure}

With two simulated sample data, we test the conditional independence between $x_1,x_2,x_3$ and $y$ given the controlling parameters $\rho$ or $\theta$ respectively. The proposed method was used for such testing. The \textsf{R} package \texttt{copent} \cite{ma2021copent,Ma2021} was used as the implementation of the CE-based CI test in the experiments. The results of the tests of the two experiments are shown in Figure \ref{fig:simresults}. It can be easily learned from it that in both experiments, $x_1,x_2$ and $y$ are conditionally dependent given the controlling parameters while $x_3$ and $y$ are conditionally independent given the same controlling parameters.

\section{Applications on real world data}
\subsection{Adult census income data}
We also applied our method on two real world data. The proposed method was used for such testing. The \textsf{R} package \texttt{copent} \cite{ma2021copent,Ma2021} was also used in the experiments. The first data is the UCI adult census income data \cite{Dua2017}. It was extracted from the 1994 US Census database and contains 32,561 records, each of which has 15 variables, including a income variable indicating whether adult’s annual income is greater than 50K dollars, and other 14 social-economic factors, such as age, education, occupation, and sex, etc. Researchers have been interested in the income inequality between males and females in the data and tried to find the causal source of this inequality. For example, Budhathoki et al. \cite{Budhathoki2021a} studied the effect of education and occupation on this incomes inequality. 

\begin{figure}
	\centering
	\includegraphics[width=0.9\textwidth]{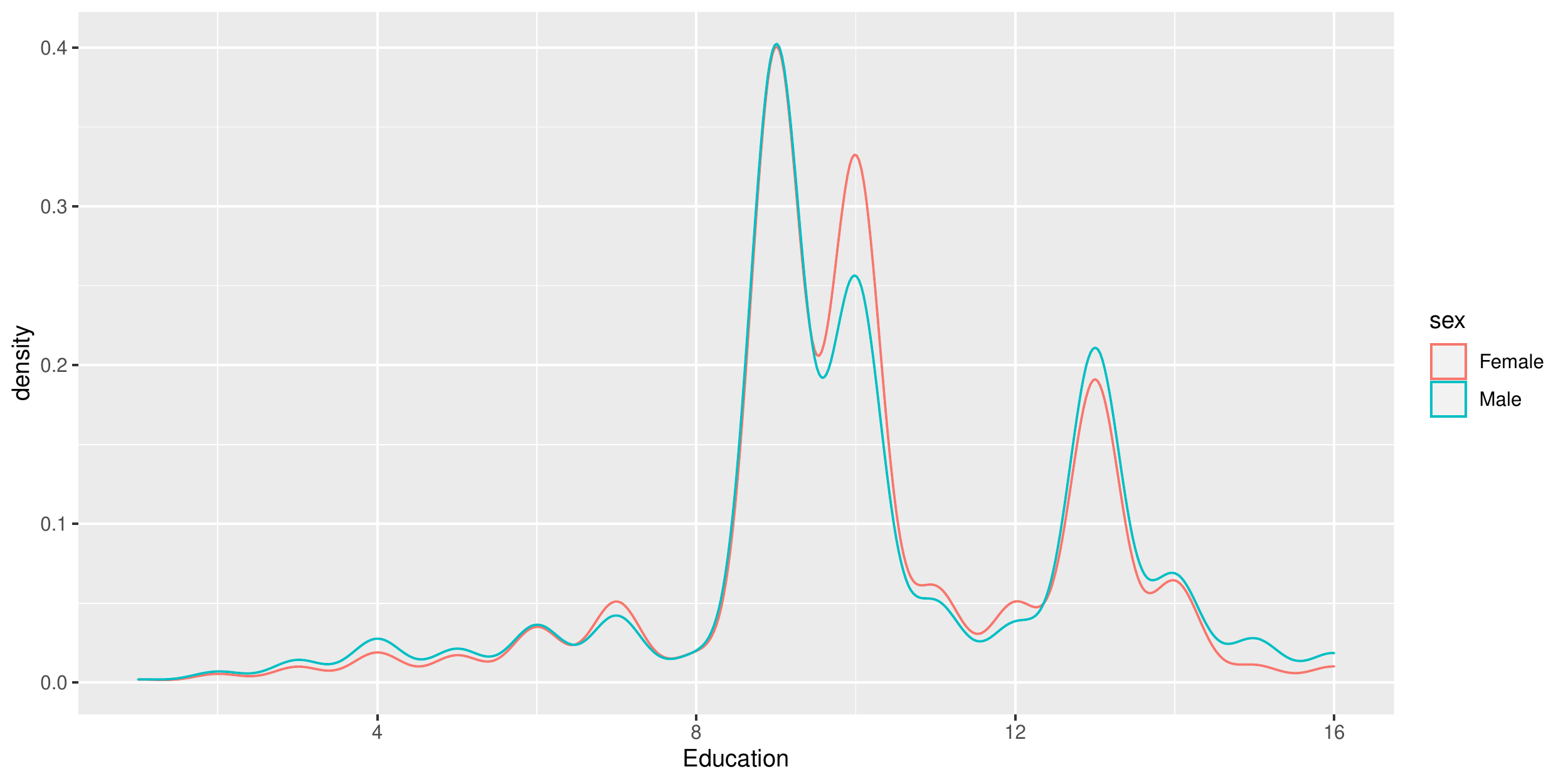}
	\caption{Distributions of education of female and male in the adult census income data.}
	\label{fig:distofedu}
\end{figure}

\begin{figure}
	\centering
	\includegraphics[width=0.65\textwidth]{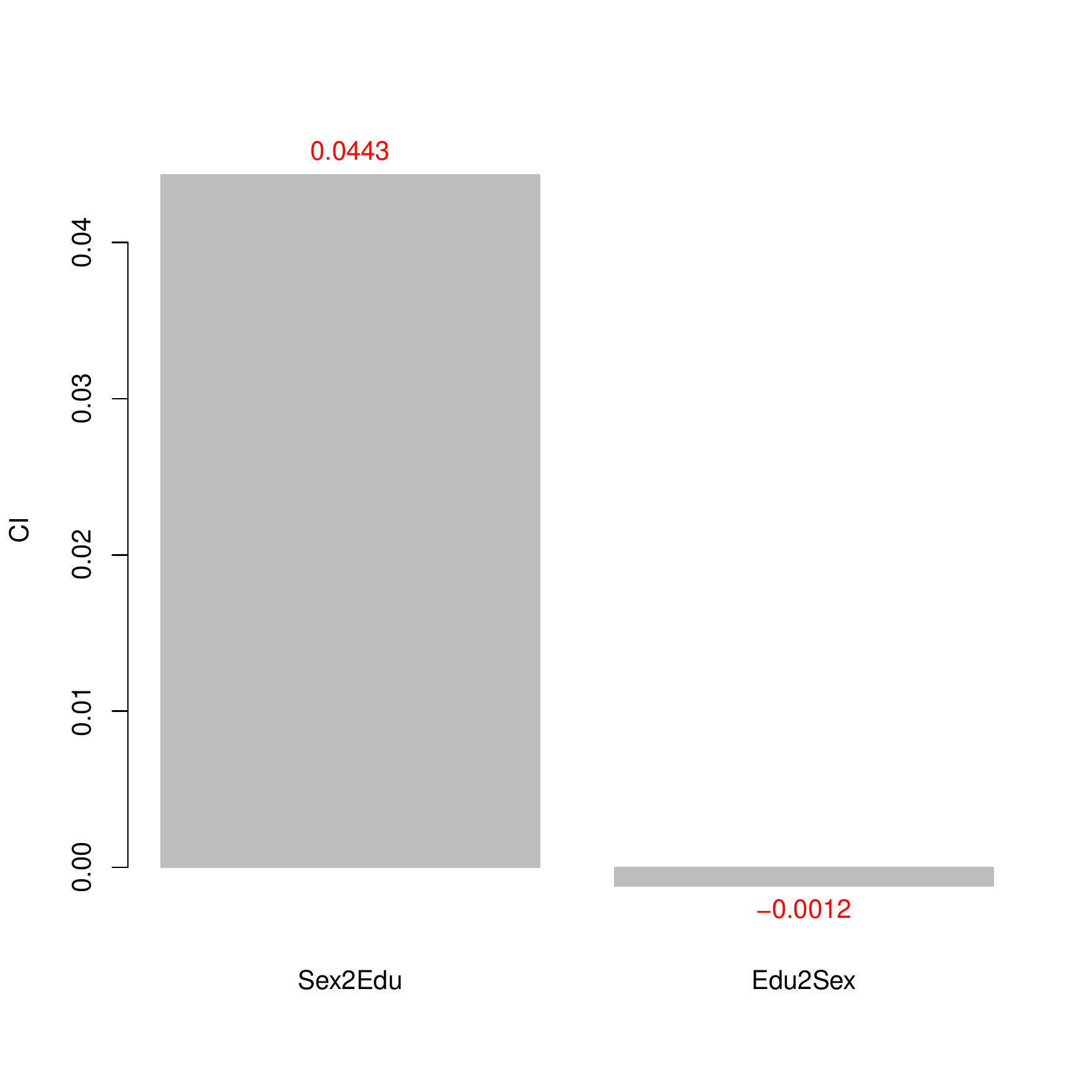}
	\caption{Results of the CI tests on the adult census income data.}
	\label{fig:adultresults}
\end{figure}

Here we consider this problem as a DA problem with sex or education factors as intervention. We hypothesis that sex cause the distribution change of education and then lead to income inequality. It can be seen from Figure \ref{fig:distofedu} that the distributions of education of female and male are different. To test the hypothesis, we applied our method to test the conditional independence between income and education given sex. As contrast, we also test the conditional independence between income and sex given education. The results of the two tests are shown in Figure \ref{fig:adultresults}. It can be learned from it that the former test presents high value of CI strength while the latter almost zero value, which implies that the first hypothesis is true instead of the second.

\subsection{Gait characteristics data}
We applied out method to the gait characteristics data in the motivation problem. The data was collected from the elderly in Tianjin and Chengdu, China, and has been studied for developing a technology for fall risk assessment \cite{ma2020predicting}. The gait characteristics data extracted from the original data reflect the functional performance of the elderly in two scenarios: TUG test and daily life. Previous study \cite{ma2022comparison} has shown the difference of gait characteristics between these two scenarios. Therefore, the data can be considered as multi-domain data. We are interested in finding the relationship between gait characteristics and TUG score that is invariant in both scenarios. 

\begin{figure}
	\centering
	\includegraphics[width=0.95\textwidth]{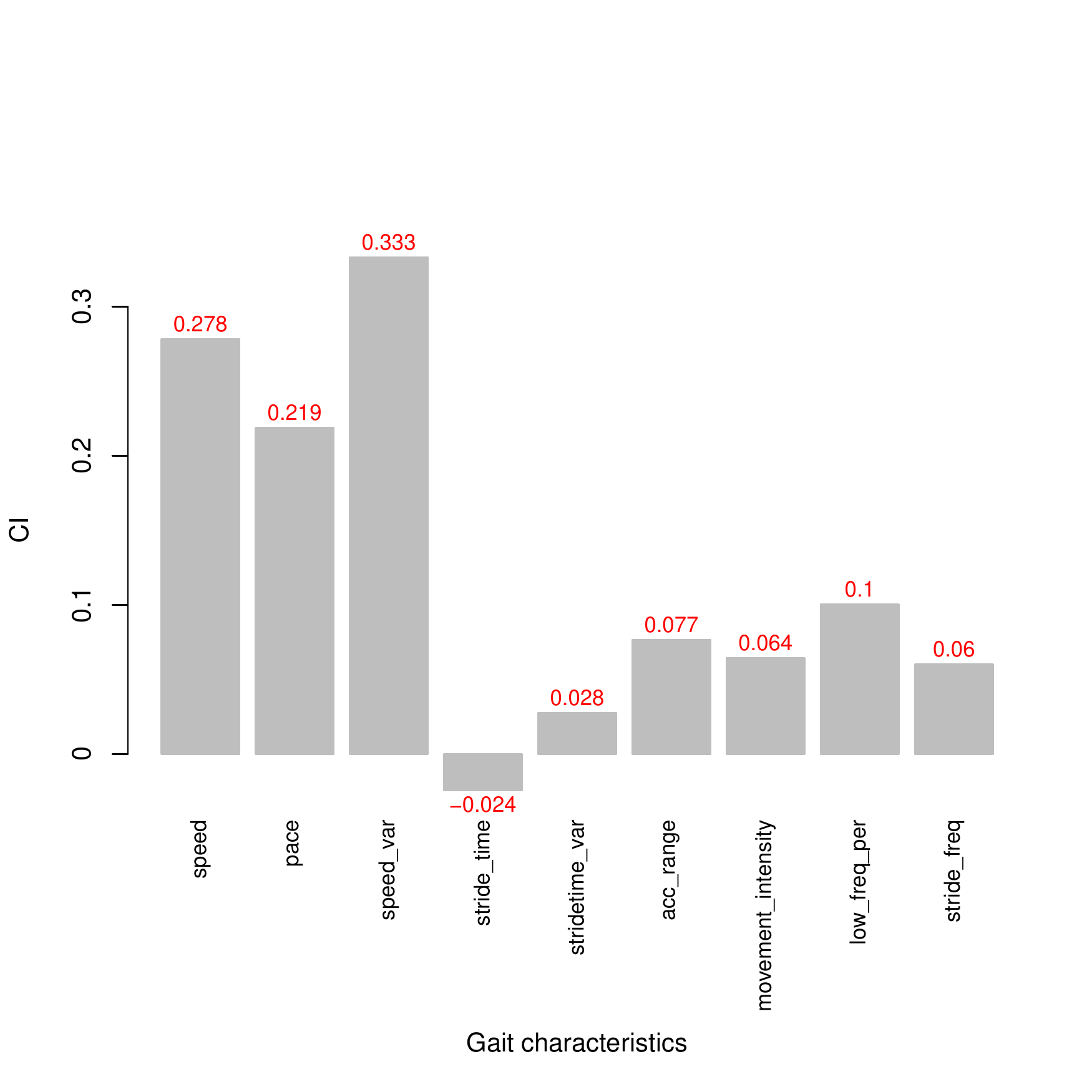}
	\caption{Results of the CI tests on the gait characteristics data.}
	\label{fig:gaitresults}
\end{figure}

In this experiment, the data of the 9 gait characteristics from two interventions (TUG test and daily life) on the target population are put together. Each gait characteristics data was attached with the corresponding TUG score. And then the intervention variable (I = 1 for TUG test, I = 2 for daily life) was augmented to the data. The CE-based CI test was performed on the augmented data to estimate the value of CI measure with the proposed method. The estimation results is shown in Figure \ref{fig:gaitresults}, from which one can learn that speed, pace, speed variability has much higher value of CI measure than the other 6 gait characteristics. This suggests that these 3 characteristics are conditionally dependent on TUG score given interventions much stronger than others.

\section{Discussion}
In this work, we proposed to use CE-based CI test to solve the causal DA problem. Our idea is similar to the previous work \cite{magliacane2018domain,Mooij2020,zhang2020domain}. We all try to tackle the problem from the view of causality. However, we use different tools for causal discovery. Conditional independence testing with CE is a information-theoretical based method. It is mathematically sound and hence advantageous to other similar tools for testing CI, such as partial correlation, conditional distance correlation \cite{wang2015conditional}, kernel-based conditional independence tests \cite{zhang2011kernel}, conditional dependence coefficient \cite{Azadkia2021}, generalised covariance measure \cite{shah2020the}, and kernel partial correlation \cite{huang2020kernel}, etc. Mooij et al. \cite{Mooij2020} applied causal structure learning algorithm to find the causal relationship. The differences between theirs and ours are that 1) they assume the prediction outcome in target domain is unknown; 2) they use partial correlation for CI test while we use the CE-based one. Zhang et al. \cite{zhang2020domain} also have a similar idea to ours. However, they tried to solve the problem by using latent-variable Generative Adversarial Network to model the underlying distributions. Oberst et al. \cite{Oberst2021} built a linear structural causal model for this problem and learned it with regularization technique for balancing performance and invariance. As contrast, our method does not assume linearity and is model-free.

Compared with the DA methods that try to find a invariant representation with dependence measures \cite{Sun2016,Chen2020,zhao2021domain,chen2022preserving,Long2015}, causal DA is more reasonable in that it can explain the domain discrepancy with a causal source outside of systems. The assumption that there is such a invariant representation is hard to verify and usually invalid in practice. Finding causality for the discrepancy between domains could be much wiser than naively searching the `invariant' correlations.

In the real-world data experiment, the three assumptions (exogeneity, randomization, and genericity) were strictly obeyed in the experiments. In adult census income data, it is clear that sex is exogenous to education, randomly assigned, and confounded with each other. In gait characteristics data, the TUG test is exogenous to the subject to be tested; the TUG test is randomly conducted on the subject irregardless of their gender, education, health conditions, or other factors. The assumption of genericity is also held in the experiment since the intervention is simply with only two values. We consider TUG test and daily life as two contexts/domains that intervene the movement state of human body and lead to the change of the (unknown) mechanism of movement. The movement states can be reflected by the gait characteristics that quantify the movement of body. The TUG score that measures functional ability is considered as a stable property of human body to be invariant to interventions. 

Under these assumptions, we test the conditional independence in two real-world data with our methods. Since the CE-based CI test is a non-parametric method without any assumption on the underlying systems, it is applicable to our problem even when we assume the unknown mechanism of the system (social system or body system) and the unknown relationships between state variables (social-economic factors or gait characteristics). In the same reason, we can also make no assumption on the invariant function from certain state variables to the observable variable (income or TUG score).

\section{Conclusions}
Domain Adaptation (DA) aims to find a transferable model on source domain to target domain. As a special case of DA, causal DA tackles the problem from the view of causality. It transforms DA as a causal discovery problem by embedding the relationships in multiple domains in a larger causal structural network and tries to find the invariant relationships and also the causal source of distribution drifts across domains. 

In this paper, we proposed a method for causal DA with the CE-based CI test that can find the invariant causal relationships across domain. We introduce the theory that see multiple domains as the effects of interventions on a system. The invariant relationships between the observable variables and the state variables of the systems given interventions can then be discovered by the CE-based CI test. We conducted two simulation experiments to evaluate the proposed method. The simulation results show the effectiveness of the method. The proposed method was then evaluated on two real-world data: the UCI adult census income and our gait characteristics data. On the first data, the proposed method found that education and income are conditionally dependence given sex. On the second data, the proposed method found that gait speed, pace, and speed variability are conditionally dependent with TUG score in two scenarios (TUG test and daily life). Both discoveries are reasonable and meaningful and hence demonstrate the power of the proposed method.

\bibliographystyle{unsrt}
\bibliography{cda}

\end{document}